\documentclass{article}

\usepackage[final]{neurips_2024}

\makeatletter
\def\@noticestring{}   % Completely empties the notice/footer
\makeatother

\usepackage[utf8]{inputenc}
\usepackage[T1]{fontenc}
\usepackage{hyperref}
\usepackage{url}
\usepackage{booktabs}
\usepackage{amsfonts}
\usepackage{amsmath}
\usepackage{nicefrac}
\usepackage{microtype}
\usepackage{xcolor}
\usepackage{graphicx}
\usepackage{float}
\usepackage{multirow}

\title{YC Bench: a Live Benchmark for Forecasting Startup
Outperformance in Y Combinator Batches}

\author{%
  Mostapha Benhenda \\
  \texttt{mostaphabenhenda@gmail.com} \\
}

\begin{document}

\maketitle

\begin{abstract}
Forecasting startup success is notoriously difficult, partly because meaningful outcomes, such as exits, large funding rounds, and sustained revenue growth, are rare and can take years to materialize. As a result, signals are sparse and evaluation cycles are slow. Y Combinator batches offer a unique mitigation: each batch comprises around 200 startups, funded simultaneously, with evaluation at Demo Day only three months later.

We introduce YC Bench, a live benchmark for forecasting early outperformance within YC batches. Using the YC W26 batch as a case study (196 startups), we measure outperformance with a Pre-Demo Day Score, a KPI combining publicly available traction signals and web visibility. This short-term metric enables rapid evaluation of forecasting models.

As a baseline, we take Google mentions prior to the YC W26 application deadline, a simple proxy for prior brand recognition, recovering 6 of 11 top performers at YC Demo Day (55\% recall). YC Bench provides a live benchmark for studying startup success forecasting, with iteration cycles measured in months rather than years. 

Code and Data are available on GitHub: 

\url{https://github.com/benstaf/ycbench}
\end{abstract}

\section{Introduction}

Forecasting startup success is a longstanding challenge in the industry. Better predictions could improve
capital allocation, accelerate innovation, and inform decision-making by
investors and founders. Improvements in forecasting are being hindered by a
fundamental constraint: meaningful indicators of success, such as exits, large funding
rounds, sustained revenue growth, are rare events that take up to 7-10 years to materialize.

This creates a challenging learning environment. Labels
are sparse, evaluation cycles are slow, and
retrospective analyses suffer from survivorship bias and market
condition confounding. As a result, unlike other domains, such as computer vision or natural
language processing, startup prediction lacks standardized, rapidly
evaluable benchmarks.

To address these challenges, we introduce YC Bench, a live benchmark for
forecasting early outperformance within Y Combinator batches. YC startup accelerator provides a unique setting: each batch consists of approximately 200
startups, funded simultaneously, operating under shared constraints, and
evaluated at a common endpoint---Demo Day--- three months later.
This enables short-term, comparable
outcome measurement, facilitating the
study of early indicators of startup performance.
We measure batch outperformance using a Pre-Demo Day Score, a metric combining
publicly available traction signals and web visibility.

\section{Related Work}

\subsection{Predicting Startup Success}

Predicting startup outcomes is attracting increasing attention across
entrepreneurship, finance, and machine learning. Studies
using structured datasets from Crunchbase, PitchBook and LinkedIn have modeled long-term outcomes such as acquisitions, IPOs, and follow-on funding rounds, using retrospective data \cite{krishna2016predicting,
hunter2018,gastaud2019varying,bonaventura2019,corea2021,jinze2021,LiuHL23, chen2025,maarouf2026}.

These methods rely on labels that take years to
resolve, making iterative development difficult. That's why YC Bench focuses on short-term outperformance within YC batches.

\subsection{Web Visibility and Online Signals}

Online signals such as web mentions, search queries, and engagement correlate with venture growth, funding outcomes, and market traction \cite{sharchilev2018web,gavrilenko2023,antretter2019predicting}. 

YC Bench adopts this perspective by using Google search mentions. Compared with other online metrics, Google mentions provide relatively consistent coverage across early-stage YC startups, suitable for a batch-wide attention signal.

\subsection{Startup Accelerator Batches as Benchmark Environments}

The effects of startup accelerators on outcomes have been examined in the literature \cite{hallen2014accelerators, cohen2019design}, but the batch structure itself has received less attention, although in practice, it is leveraged by YC-specific Venture Capital funds for their investments  (Rebel Fund, Lobster Capital...). YC Bench is, to our knowledge, the first benchmark to explicitly leverage the batch structure of a startup accelerator.

\subsection{Live Forecasting Benchmarks}

Live forecasting benchmarks allow to avoid contamination from training data, because their tasks are to predict the future. For example, FutureX \cite{zeng2025futurex} continuously aggregates prediction questions across politics, finance, and technology, scoring models once outcomes resolve. FutureX-Pro \cite{liu2026futurexpro} extends this to high-stakes verticals including finance and public health. ForecastBench \cite{forecastbench2024} evaluates models against human forecasters on prediction-market-style questions; Metaculus \cite{metaculus2024} hosts large-scale forecasting tournaments. OpenForecaster \cite{chandak2025openforecaster} trains specialized forecasting models on automatically generated news-derived prediction datasets.

YC Bench applies this forward-looking approach to the narrow setting of YC batches, reducing cross-domain heterogeneity.

\subsection{Bridging Short-Term Signals and Long-Term Outcomes}

When long-term outcomes are delayed, many domains use short-term proxy signals. In medical research, biomarker levels can be used as surrogate endpoints in clinical trials, instead of long-term survival outcomes \cite{prentice1989surrogate, fleming1994surrogate}. That's especially the case for diseases like Alzheimer's, where clinical outcomes can take decades to resolve \cite{chen2026amyloid}. In information retrieval, implicit feedback signals such as clicks and dwell time serve as proxies for relevance judgments \cite{joachims2002optimizing}.

Startup forecasting faces the same challenge: meaningful outcomes, such as exits or survival, take 7--10 years to resolve, making them impractical as near-term training targets. That's why YC Bench introduces a Pre-Demo Day Score as a short-term observable proxy for long-term outperformance.

\section{YC Bench Benchmark Design}

\subsection{Prediction Task}

The prediction task is to identify the top-10\% of startups of the batch, as they appear the day before YC Demo Day.
Top-10\% YC startups drive 96\% of all investor returns. This share corresponds to the best risk-adjusted YC portfolio strategy, according to YC-specific Rebel Fund investor \cite{heyman2024powerlaw,heyman2025bullseye}.

YC Bench participants should ideally post their predictions right after the publication of the new batch composition by Y Combinator.
The idea is that to access the best YC deals, startup investors need to identify them earlier than the crowd at Demo Day.

\subsection{Evaluation Metrics of Predictive Models}

Predictions are evaluated against observed Pre-Demo Day Scores, using
two complementary metrics:

\begin{itemize}
  \item \textbf{Precision@20}: fraction of the top-20 predicted startups that appear among the 20 highest Pre-Demo Day Scores (top 10\% of the batch).

  \item \textbf{Recall@11}: fraction of the 11 high-traction startups recovered within the top-20 predictions. This traction data was disclosed in \cite{lobstercap2026ycw26} and on LinkedIn for the W26 batch. The theoretical maximum is 20, corresponding to the full top 10\% of the batch.
\end{itemize}

\section{The Pre-Demo Day Score}

The Pre-Demo Day Score combines two complementary
signals: a traction score for startups with disclosed performance data,
and an attention score for the remainder.

\subsection{Traction Score}

The traction score uses a max-of-weighted-metrics
aggregation, which is robust to the sparse reporting typical of early-stage startups:

\begin{equation}
  S_{\text{traction}} = \max_{k \in \{ \text{signals} \}} \left( w_k \cdot x_k \right)
\end{equation}

where $x_k$ is the value of signal $k$ and $w_k$ is its weight.
Table~\ref{tab:weights} lists the signals and weights used in the W26
instance: revenue signals
dominate, engagement follows, and acquisition signals are discounted. These weights are heuristically instantiated in this initial benchmark, but will be learnt from data in future iterations.

\begin{table}[h]
  \centering
  \caption{VC signal weights for the traction score.}
  \label{tab:weights}
  \begin{tabular}{llp{6cm}}
    \toprule
    Signal & Weight & Rationale \\
    \midrule
    ARR / Revenue          & 1.00 & Strongest signal of product-market fit \\
    Pilot Revenue          & 0.50 & Confirmed commercial interest \\
    LOI / Signed Contracts & 0.20 & Intent to pay, not yet revenue \\
    Active Users           & 0.40 & Engagement at scale \\
    Activity Volume        & 0.25 & Usage depth proxy \\
    Signups                & 0.15 & Acquisition funnel entry \\
    Ecosystem Pull         & 0.10 & Developer adoption signal \\
    \bottomrule
  \end{tabular}
\end{table}

When month-over-month growth is available, a velocity multiplier
captures momentum:

\begin{equation}
  S_{\text{traction}} := S_{\text{traction}} \cdot
  \left(1 + 10 \cdot g_{\text{MoM}} \right)
\end{equation}

The coefficient 10 rewards growth: a 50\%
month-over-month growth rate, like Pocket (YC W26) reported, yields a $6\times$ multiplier.

\subsection{Attention Score}

For startups without traction data---the majority of any batch---we
compute an attention score from Google web mention counts:

\begin{equation}
  S_{\text{attention}} = w_m \cdot x_m
\end{equation}

where $x_m$ is the number of Google search mentions for the startup's
domain during the YC batch, and $w_m =0.05$ (see Section~\ref{sec:dataset}). We focus on Google mentions as the primary attention signal because
alternatives---Google Trends, GitHub stars, App Store rankings, LinkedIn
headcount, Hacker News, Product Hunt---are too sparse across almost all startups of the batch.

\subsection{Pre-Demo Day Score}

The Pre-Demo Day Score combines both signals via max aggregation:

\begin{equation}
  \text{Pre-Demo Day Score} = \max\!\left(S_{\text{traction}},\;
  S_{\text{attention}}\right)
\end{equation}

This ensures high-traction startups rank highly even without broad web
visibility, while high-visibility startups are captured even without
disclosed traction.

\begin{figure}[H]
\centering
\includegraphics[width=0.85\textwidth]{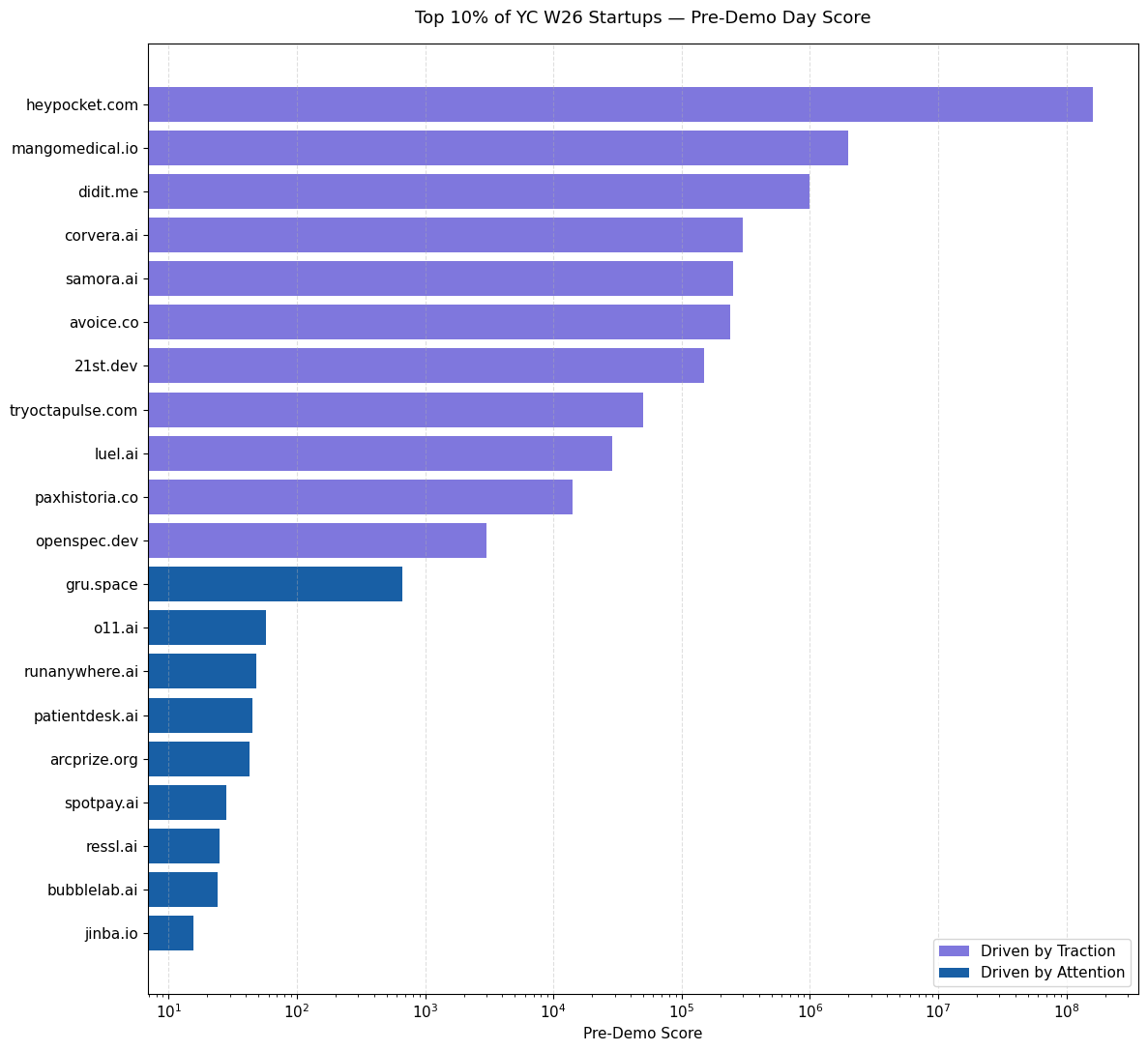}
\caption{Top 10\% of YC W26 startups by Pre-Demo Day Score.
Purple = driven primarily by traction,
Dark blue = driven primarily by attention (Google mentions).}
\label{fig:predemodayscore}
\end{figure}

\section{The YC W26 Dataset}
\label{sec:dataset}

\subsection{Traction Data}

Traction data for 11 startups was sourced from a publicly circulated analysis by Lobster Capital~\cite{lobstercap2026ycw26} and from self-disclosures on LinkedIn. Signals covered include ARR, pilot revenue, letters of intent, active users, and activity volume.

\begin{figure}[H]
\centering
\includegraphics[width=0.85\textwidth]{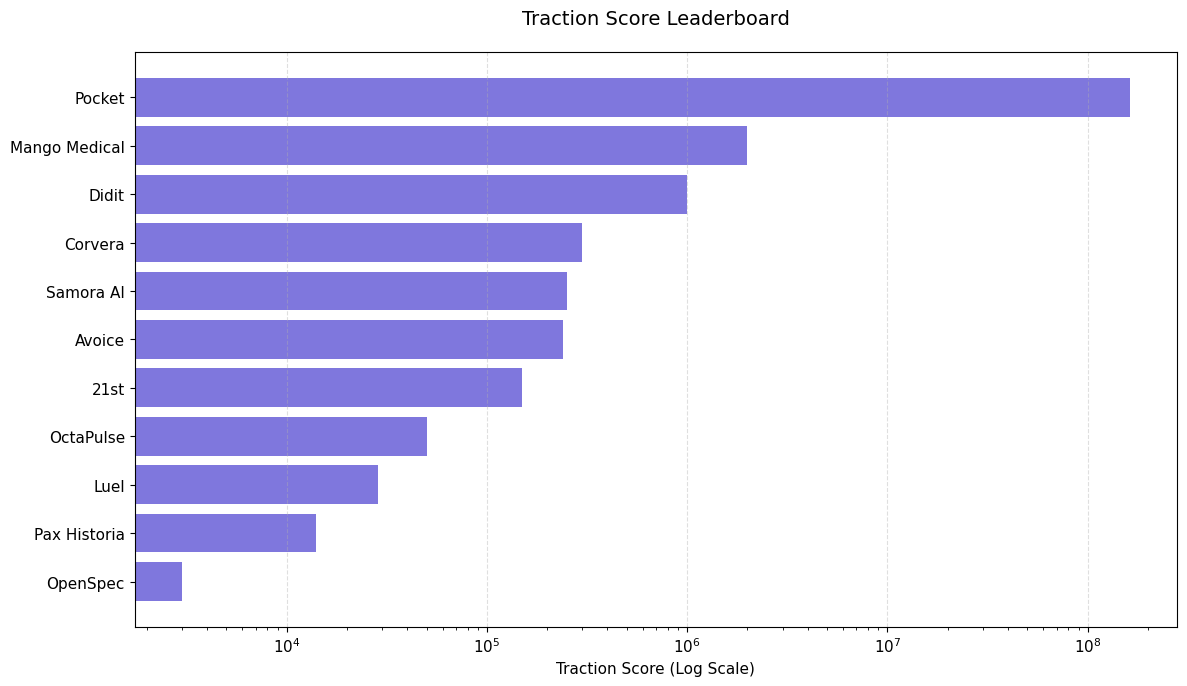}
\caption{Traction Score Leaderboard for YC W26 (log scale).}
\label{fig:traction_leaderboard}
\end{figure}

\subsection{Attention Data: YC Startups Domain Mentions on Google}

Domain names were collected from the public YC company directory.
Google mention counts were retrieved via SerpAPI, querying each startup's domain from January 1st 2026 to March 17th 2026. Queries matching common expressions, or homonyms with established brands, were filtered out, yielding 184 startups with detectable web presence.

\begin{figure}[H]
\centering
\includegraphics[width=0.85\textwidth]{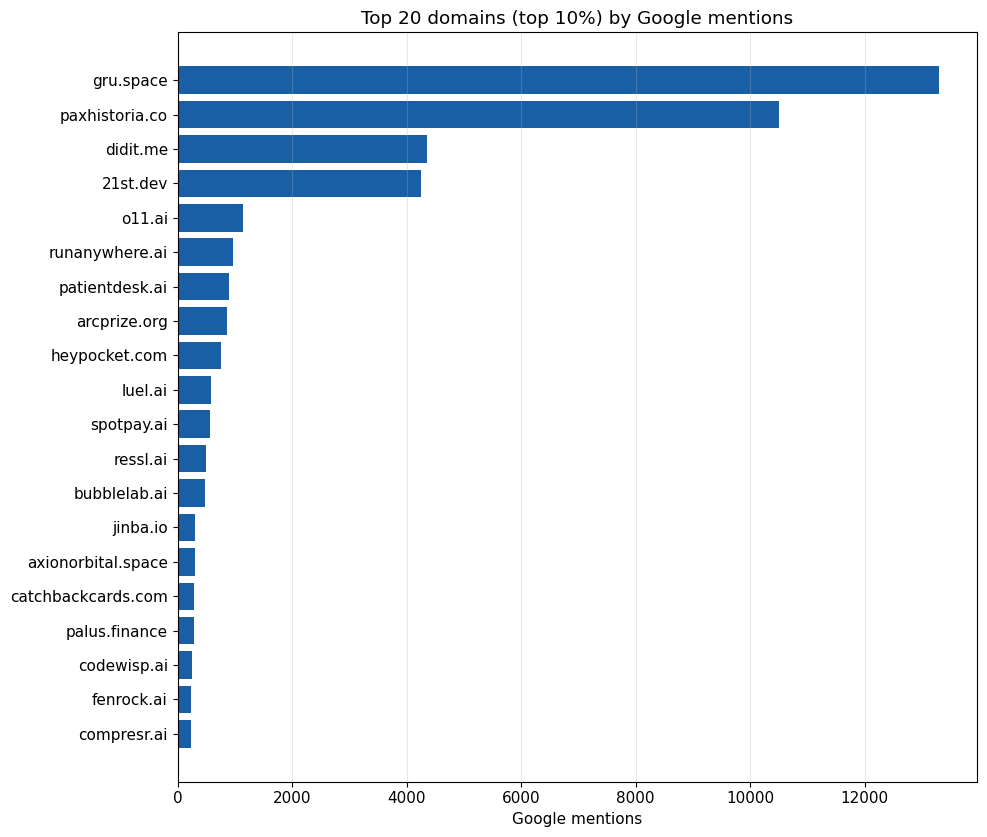}
\caption{Top 20 domains (top 10\%) by Google mentions during the YC W26 batch
(January 1 -- March 17, 2026).}
\label{fig:googlementions_post}
\end{figure}

\subsection{Attention Distribution}

Google mentions are highly concentrated, with a Gini coefficient of 0.85. The top 10\% (20 startups) accounts for around 82\% of total
mentions; the top 2\% (4 companies) accounts for over 50\%. They span various sectors---space technology, gaming, identity infrastructure, and developer tools---
showing that pre-application visibility is not sector-specific.

We heuristically set $w_m = 0.05$, so that a startup
at the median of the top-10\% attention tier (${\sim}500$ mentions)
receives a Pre-Demo Day Score comparable to a startup reporting \$25K in
pilot revenue.

\begin{figure}[H]
\centering
\includegraphics[width=\textwidth]{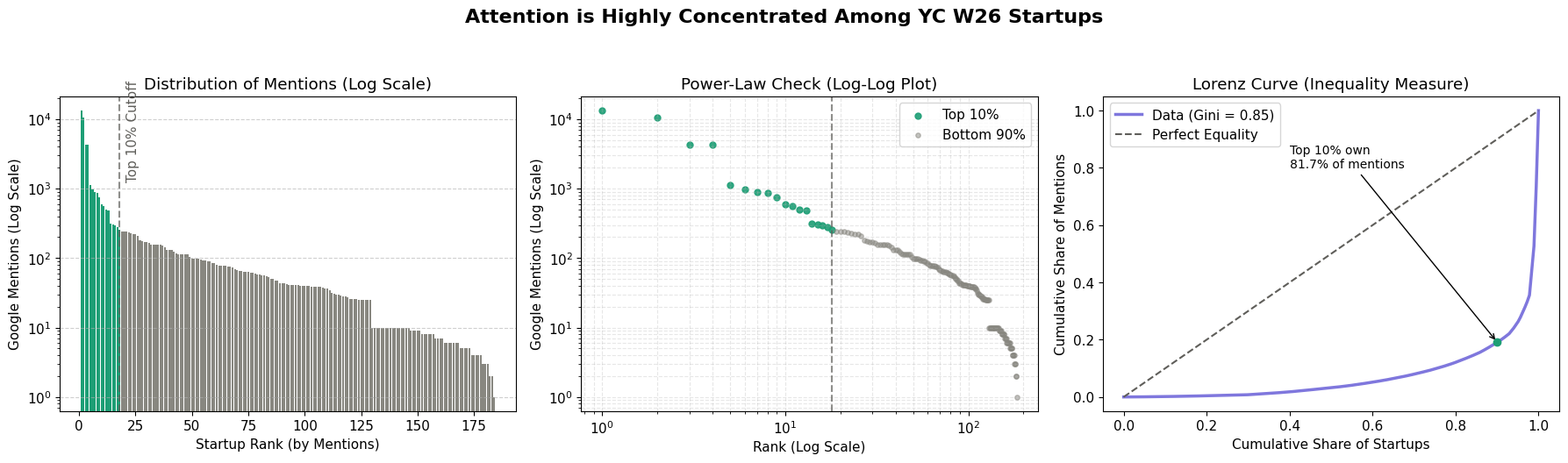}
\caption{Attention is highly concentrated among YC W26 startups.
Left: Distribution of Google mentions (log scale).
Middle: Power-law check (log-log plot).
Right: Lorenz curve showing extreme inequality (Gini = 0.85).
The top 10\% of startups capture 81.7\% of all mentions.}
\label{fig:powerlaw}
\end{figure}

\section{Baseline: Google mentions before YC W26 application deadline}

\subsection{Baseline Definition}

As a simple baseline, we use the number of Google search results
mentioning a startup's domain during a pre-application window of 67
days (a similar length to the post-application window), from August 17
to October 31, 2025, ending before the YC W26 application deadline of
November 10. This signal captures pre-existing visibility and brand
recognition prior to any YC-related exposure.

We rank startups by pre-YC application deadline mention count, removing common expressions or homonyms as previously. We select the top-20, and we evaluate this baseline against the top-20 startups by Pre-Demo Day Score.

\begin{figure}[H]
\centering
\includegraphics[width=0.85\textwidth]{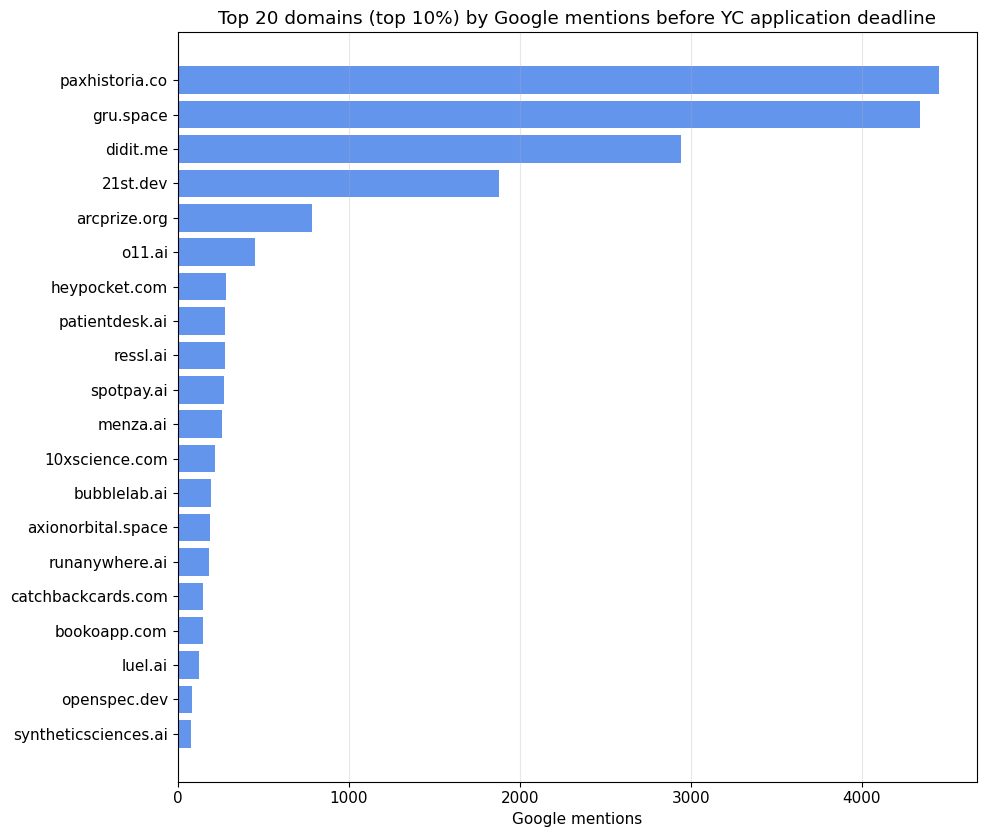}
\caption{Top 20 domains by Google mentions \emph{before} the YC W26
application deadline (August 17 -- October 31, 2025). This is the signal used by the baseline.}
\label{fig:googlementions_pre}
\end{figure}

\subsection{Baseline Performance}

\begin{table}[H]
  \centering
  \caption{Baseline predictor results on YC W26. The random predictor
  assumes uniform probability over startups.}
  \label{tab:results}
  \begin{tabular}{lll}
    \toprule
    Metric               & Baseline       & Random \\
    \midrule
    Precision@20         & 70\%         & 10\% \\
    Recall@11            & 55\%         & 10\%     \\
    Lift over random     & $7\times$   &  $1\times$     \\
    Forecasting horizon  & ${\sim}5$ months & ---  \\
    \bottomrule
  \end{tabular}
\end{table}

The baseline recovers 6 of the 11 high-traction startups (Recall@11 = 
55\%) and places 14 of 20 correctly in the top-20 predicted set 
(Precision@20 = 70\%), representing a 7$\times$ lift over random at 
a forecasting horizon of approximately five months.

The 5 missed high-traction performers were operating in narrow B2B 
verticals with limited public-facing web presence on Google.

\begin{figure}[H]
\centering
\includegraphics[width=0.85\textwidth]{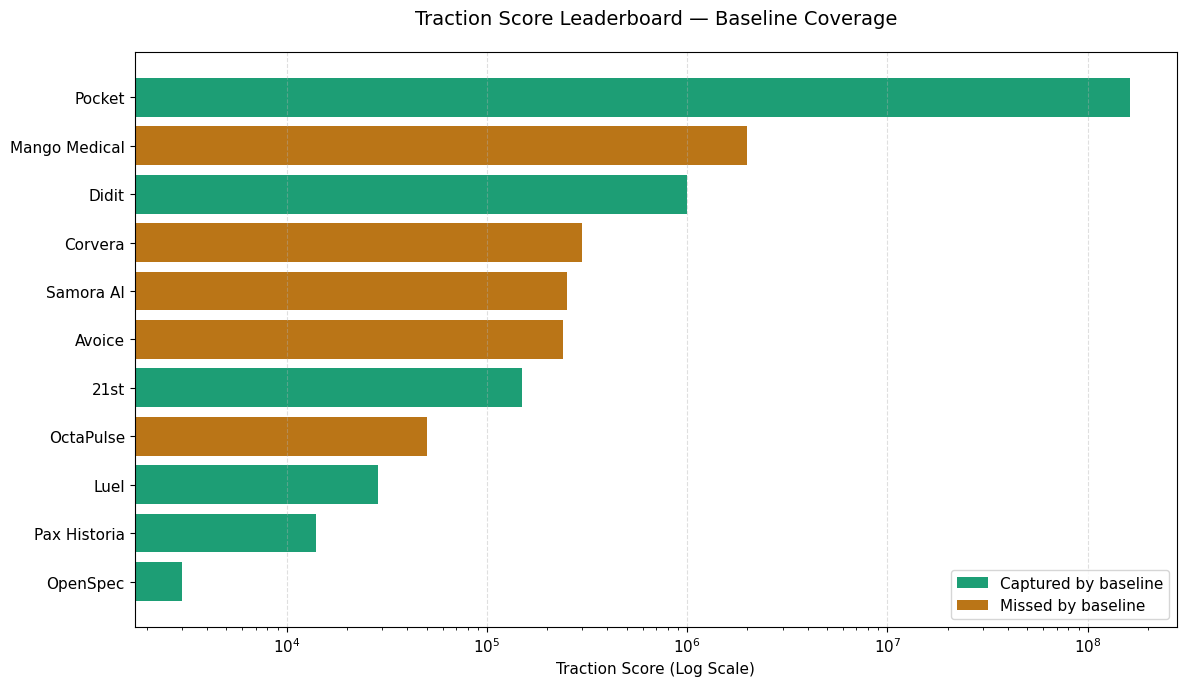}
\caption{Traction Score Leaderboard --- Baseline Coverage. Green = captured by the Google-mentions baseline,
Orange = missed by the baseline.}
\label{fig:traction_baseline}
\end{figure}

\begin{figure}[H]
\centering
\includegraphics[width=0.85\textwidth]{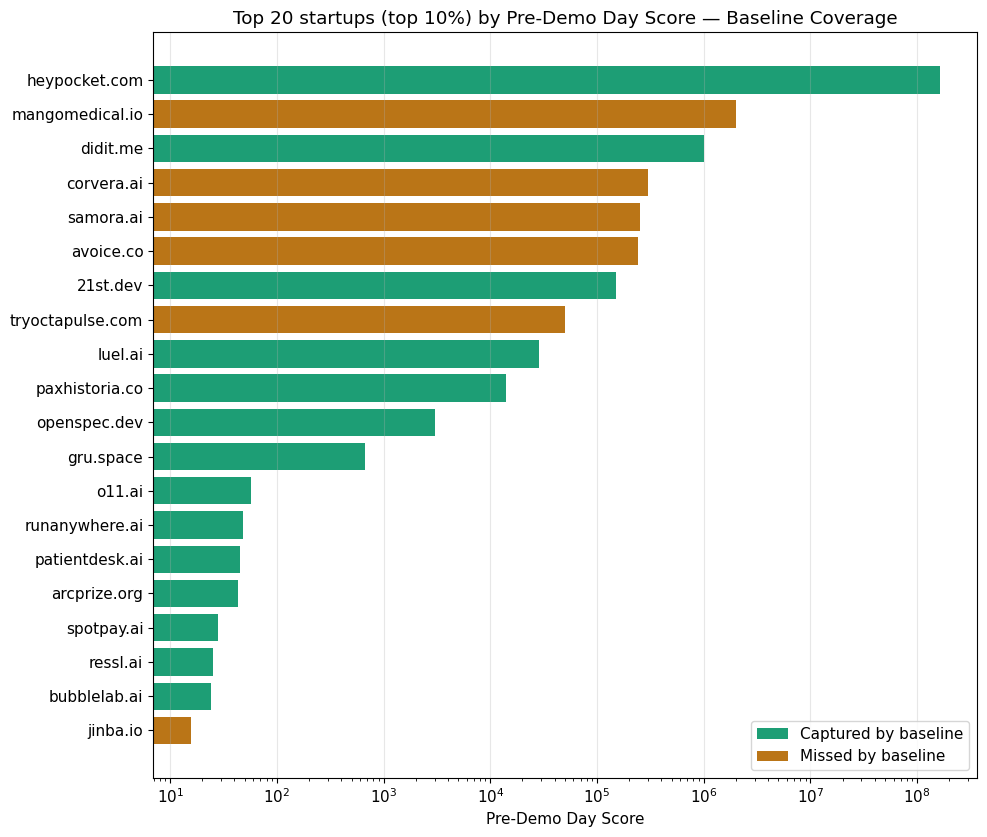}
\caption{Top 20 startups by Pre-Demo Day Score --- Baseline Coverage.
Green bars = captured by the pre-application Google mentions baseline,
Orange = missed.}
\label{fig:predemo_baseline}
\end{figure}

\clearpage

\section{Future Work}

YC Bench is designed as an evolving benchmark, and several extensions
can strengthen its reliability and scope. First, the current Pre-Demo Day Score relies on heuristic weights for
combining traction and attention signals. While these coefficients reflect
common venture capital heuristics, they are not empirically calibrated. The next step is to learn these weights from historical batches, by optimizing the
correlation between early signals and long-term outcomes.

Second, traction data in the W26 instance is available for only 11
startups. Expanding traction data collection
across future YC batches will improve the benchmark.

\section{Conclusion}

Forecasting startup success is challenging because meaningful outcomes
are rare and slow to resolve. This limits the ability to iteratively
develop predictive models. YC Bench addresses this
constraint by introducing a short-term proxy outcome---the Pre-Demo Day
Score---that can be evaluated within a few months of batch formation.

Using the YC W26 batch as a case study, we show that even a simple
baseline based on Google mentions before the YC application deadline can recover a
substantial fraction of the highest-traction startups months in
advance. This shows that publicly observable signals contain substantial information about YC startups batch outperformance.

YC Bench provides a forward-looking benchmark for studying
startup forecasting, enabling evaluation cycles measured in months rather
than years.

\bibliography{ycbench}
\bibliographystyle{plainnat}

\end{document}